\documentclass[11pt]{article}
\usepackage[margin=0.82in]{geometry}
\usepackage{amsmath,amssymb}
\usepackage{booktabs}
\usepackage{tabularx}
\usepackage{array}
\usepackage{graphicx}
\usepackage{microtype}
\usepackage{enumitem}
\usepackage[round,authoryear]{natbib}
\usepackage[hidelinks]{hyperref}
\usepackage{caption}
\usepackage{xcolor}
\usepackage{float}
\usepackage{multirow}
\usepackage{makecell}
\usepackage{url}

\setlist{nosep,leftmargin=*}

\newcolumntype{Y}{>{\raggedright\arraybackslash}X}
\newcolumntype{C}[1]{>{\centering\arraybackslash}p{#1}}
\newcolumntype{L}[1]{>{\raggedright\arraybackslash}p{#1}}

\title{\textbf{Context by Distinct Information:}\\
An Auditable Dirichlet-Process Working Memory\\
for Long, Redundant Context Streams}
\author{Siddharth Pal \and Viktoria Rojkova}
\date{Preprint}

\begin{document}
\maketitle

\begin{abstract}
Context engineering decides what information a model carries forward---conversation turns, retrieved passages, tool results, observations, and intermediate state---and current designs meter that context in tokens, whether by compressing the past into a bounded recurrent state, keeping a key--value entry for every token, or imposing a fixed budget through a window or eviction rule. All three make the token the unit of memory even when the stream is redundant and the task depends on the distinct information it carries. Building on a companion mechanism paper that opens a cache slot only when an incoming key is novel, so that memory scales with the number of distinct items rather than tokens, we develop that allocate-on-novelty cache as a working-memory component and organize context by how a task depends on the past: recall-carried information, which a later query may need from a particular earlier item, belongs in a content-addressed novelty cache; summary-carried information belongs in a recurrent state; and locality-carried information belongs in a recency window.

The claim is empirical and bounded. On a matched character-level control, novelty-gated attention reaches the performance of full attention while attending to about half the tokens, and coupling the cache with a state-space summary matches full-attention coupling at that reduced cost; retaining full attention beside the cache improves the read further still, so the novelty-kept view supplies signal rather than merely approximating attention. The advantage grows as context lengthens, while a sliding window remains preferable on short, locality-dominated spans. The same pattern holds end to end on real structured context: on next-code prediction over synthetic Medicare claims the coupled component leads full attention and every fixed-budget eviction policy by roughly a third of a bit per event at a thousand-event horizon, whereas cost forecasting over the same stream is summary-carried and the cache is neutral; and which recurrent summary to pair with the cache is itself domain-dependent, since an input-gated variant that wins on text collapses on the structured stream. Across log, clinical, claims, and mapping streams the retained memory is an inspectable table of templates, codes, drugs, or places rather than an opaque state, though a distribution shift in one log stream defeats every read and scopes the claim. The experiments are small-scale, use only public data, and do not yet evaluate multi-turn assistants or retrieval-augmented agents; what they establish is the architectural primitive, that context can scale with the distinct information a task may need rather than with tokens, in a working memory that is both content-addressable and auditable.
\end{abstract}

\section{The problem: context is metered in tokens, but information is not}

Context engineering is the construction and maintenance of the information exposed to a model at inference time. A context stream may contain natural-language turns, document passages, tool outputs, program state, observations, structured records, or mixtures of them. The stream can be long, hierarchical, and multi-rate: a slow-moving fact or entity can remain relevant across thousands of fast local updates. It is also usually redundant. A name recurs across a conversation, a retrieved passage restates an earlier fact, the same tool emits repeated status records, a code or identifier appears hundreds of times, and a log template fires thousands of times.

The standard memory choices nevertheless scale with tokens. A fixed-state recurrent or state-space model processes the stream in linear time and constant-size inference memory, but folds all prior information into a bounded state. Its content-addressed recall degrades once the number of independently retrievable items exceeds the state capacity, the regime isolated theoretically by copying separations and empirically by recall benchmarks for efficient sequence models \citep{jelassi2024,arora2023}. Full attention removes that cap by retaining a key--value entry for every token, but the cache grows with the stream and the compute of a dense read grows quadratically during training. Fixed windows and eviction caches bound the cost, but they decide what survives by recency, fixed roles, or attention statistics rather than by whether the information is already represented.

These are three answers to the same question: what is the unit of context? If the unit is a token, then repeated evidence is charged repeatedly. If the unit is a distinct item represented in the model's key space, a long redundant stream can be carried by a much smaller working memory. This paper studies the latter design and, more importantly, the boundary around it. Not every task needs the same past. The experiments separate three regimes:

\begin{itemize}
    \item \textbf{Recall-carried information}: a later query depends on a particular earlier item, including a rare one. A content-addressed distinct-item cache is the natural memory.
    \item \textbf{Summary-carried information}: the target depends on a long-range aggregate or smooth history. A recurrent or state-space path is the natural memory.
    \item \textbf{Locality-carried information}: the target depends mainly on the immediate past. A recency window is the natural memory.
\end{itemize}

The contribution is not a claim that novelty is universally superior. It is a context decomposition that separates recall-carried, summary-carried, and locality-carried information and assigns each to a distinct memory structure, together with trained end-to-end evidence locating where each one pays and where it does not, so that the novelty-gated cache is placed as one component of a working memory rather than a universal compressor.

\section{Theoretical anchor: remembering distinct items, not tokens}

The companion mechanism paper \citep{palrojkova2026} places an allocate-on-novelty cache between fixed-state recurrence and full attention. We use that result as the theoretical anchor and summarize only the parts needed to make the present paper self-contained.

\subsection{The Dirichlet-process allocation rule as a memory operator}

Let the cache contain occupied slots $i$, each with key $\kappa_i$, value $\nu_i$, usage count, and provenance. For an incoming key--value pair $(k_t,v_t)$, novelty is
\begin{equation}
\mathrm{nov}_t = 1 - \max_{i\in\mathrm{slots}} \mathrm{sim}(k_t,\kappa_i),
\end{equation}
where the experiments use cosine similarity. The allocation rule is
\begin{equation}
\begin{cases}
\text{append }(k_t,v_t), & \mathrm{nov}_t > \tau,\\
\text{merge into }\arg\max_i \mathrm{sim}(k_t,\kappa_i), & \text{otherwise.}
\end{cases}
\end{equation}
This is the DP-means rule, the small-variance maximum-a-posteriori limit of a Dirichlet-process mixture \citep{kulis2012}, used here not for offline latent-variable inference but as the write operator of the working memory. Repeats merge into the slot they already spawned; genuinely new items open a slot. On streams generated from a finite set, the slot count therefore saturates near the number of distinct items. A query $q$ reads over occupied slots only,
\begin{equation}
\hat v = \sum_i \mathrm{softmax}_i\!\left(\frac{\mathrm{sim}(q,\kappa_i)}{\theta}\right)\nu_i,
\end{equation}
so the read cost follows the number of slots rather than the number of observed tokens.

The companion paper establishes the mechanism-level claims that the present work relies on: at fourfold redundancy, the cache matches full-attention associative recall with a fourfold smaller cache; it dominates recency, heavy-hitter \citep{zhang2023}, sink-plus-window \citep{xiao2024}, and SnapKV-style \citep{li2024} fixed-budget policies on the recall--size frontier; beside a state-space backbone it answers both an item-recall query and a long-range aggregate at the lowest memory tested; and a minimal two-parameter novelty-threshold gate trained on task loss recovers the allocation rule exactly, whereas an over-parameterized saliency gate fails. The relevant inductive bias is novelty, not gate capacity \citep{palrojkova2026}.

\subsection{Static and surprise-adaptive context}

The base cache uses a fixed threshold $\tau$. The companion paper also defines a surprise-adaptive form for non-stationary streams. If $o_t\in\{0,1\}$ records whether step $t$ opened a slot, an exponential moving average
\begin{equation}
T_t=(1-\eta)T_{t-1}+\eta o_t
\end{equation}
tracks the recent allocation rate. The effective budget is
\begin{equation}
M_t=M_0+\beta T_t,
\end{equation}
and least-used slots are evicted whenever the occupied count exceeds $M_t$. The cache heats during bursts of novel information and consolidates when the stream settles. At matched average cost it improves recall when the working set varies, but correctly offers no advantage when the demand is stationary \citep{palrojkova2026}. In the trained windowed models below, an explicit adaptive schedule turns out to be redundant because the learned soft gate already adapts token by token; that negative result is part of the present paper.

\subsection{Why this is a context-engineering primitive}

The cache changes context construction in three ways.

First, it changes the scaling variable from observed tokens to occupied novelty classes. The claim is operational, not philosophical: the cache size is the admitted-slot count, and all reported savings use that measured count.

Second, it makes the retained context inspectable when the frontend gives slots recognizable semantics. A slot can expose its first occurrence, merged occurrences, usage, key, value, and source offsets. An eviction cache can report what survived, but not why two repeated observations became one object; a fixed-state model cannot expose a discrete retained set at all. The auditability claim is therefore strongest in the structured streams below, where the slot table literally becomes a template catalog, formulary, code set, or place map. It is not guaranteed merely by using the gate: interpretability depends on the key representation and merge policy.

Third, it separates context construction from context reading. Allocation can be more expensive than appending every token, but repeated reads operate on the much smaller occupied set. That build-versus-read trade is appropriate when context is long-lived, queried repeatedly, or memory-bound, and inappropriate for a single read over a short sequence.

\section{Position: context engineering already allocates on novelty by hand}

The mechanism is new as a learned key--value memory for a deep sequence model, but the allocation principle is not. Several mature systems already construct context by keeping a representative only when the incoming observation is sufficiently new.

\textbf{Online log parsing.} Drain, Spell, and related production parsers stream log lines, compare each line to templates retained so far, open a template when the line is dissimilar to all of them, and otherwise merge it into the nearest template while generalizing mismatched positions \citep{du2016,he2017}. That is allocate-on-novelty with a domain-specific similarity and merge. The DP cache expresses the same loop as a differentiable memory whose threshold, similarity, and merge can in principle be learned.

\textbf{Macro-tokenization.} Long structured records are commonly aggregated into one token per claim, admission, processing step, or other schema-defined unit. A macro-token is a hand-authored slot: context is allocated per process unit instead of per raw event. The cache replaces the schema-fixed allocation with a content-driven one and can apply the same rule across levels of a hierarchy.

\textbf{Keyframe selection.} Visual SLAM systems in the ORB-SLAM line insert a keyframe when the current view is sufficiently novel against retained keyframes and close loops through nearest-neighbour lookup over that store \citep{galvez2012,murartal2015}. The memory grows with the explored environment rather than the length of the trajectory. Recent feed-forward streaming reconstruction instead compresses the past into fixed roles such as an anchor, a recency window, and a trajectory summary \citep{gct2026}; this is well suited to locality-carried geometry but cannot answer a revisit query once the relevant place has left the fixed window.

These systems are not presented as baselines for a universal context engine. They establish that novelty allocation is already a useful context-construction pattern wherever repeated observations refer to a finite or slowly growing set of objects. The paper's question is whether the same principle survives end-to-end representation learning and how it should be combined with recurrence and recency.

\section{The working-memory component}

\subsection{Backbone and reads}

The sequence model is a standard residual stack: token embedding, $\ell$ blocks, and a linear head. Within each block the mixing operator, or read, is one of:

\begin{enumerate}
    \item a diagonal state-space convolution, implemented as a learned per-channel decay applied causally, representing the fixed-state endpoint \citep{gu2022};
    \item full causal attention, representing the growing token-cache endpoint;
    \item novelty-gated causal attention, in which a token enters the key--value cache only if its key is dissimilar to every earlier admitted key, and later queries attend over admitted slots alone.
\end{enumerate}

The admitted fraction is the number of cache slots divided by the number of input tokens and is measured during training. A coupled read adds a recurrent path beside the content-addressed path inside the block: the recurrence integrates; the cache recalls. We measure four coupled forms: state-space plus full attention, state-space plus novelty-gated cache (the deployment component), all three paths together, and a selective input-gated recurrence replacing the plain state-space path \citep{gu2023}.

This architecture should be read as a context decomposition rather than a single selector. The recurrent path carries compressed history; the gated cache carries distinct addressable items; a windowed baseline carries immediate local context. The experiments determine which path matters for which target.

\subsection{What counts as a context item}

The gate operates on keys, not on an externally guaranteed ontology. A context item is therefore the unit emitted by the frontend and separated by the learned key geometry. In the control studies it is a character token. In the structured studies it is an event with categorical and hierarchical features. In a future multi-turn or retrieval-augmented system it could be a turn, sentence, chunk, tool result, entity mention, or explicit fact, but those frontends are not evaluated here and no claim is made that cosine novelty alone will discover those units without representation learning and provenance-aware merging.

For structured events, the frontend encodes two fused streams: a categorical stream containing event type, code, and code system, and a numeric stream containing amounts and durations. Feature-wise modulation fuses the streams \citep{perez2018}. Explicit boundary tokens mark hierarchy levels such as line, claim, and admission and provide natural cache segmentation points. The claims next-code study uses the categorical stream with boundary tokens. The redesigned cost task adds the larger outpatient corpus and regularization; the original plan for the numeric fusion remains a broader frontend direction. The character-level control uses no domain frontend so the read comparison is clean.

\subsection{Auditable working memory}

Each retained slot can be represented as
\begin{equation*}
(\text{key},\ \text{value},\ \text{first position},\ \text{merged positions},\ \text{usage},\ \text{source provenance}).
\end{equation*}
The model still learns the key space, so a slot is not automatically a human concept. In the structured measurements, however, the slots align with explicit domain objects. The question ``what context did the model retain?'' can then be answered by the slot table itself. This is a separate design objective from cache reduction and motivates retaining provenance even when only the key and value are needed for prediction.

\section{Mechanism evidence on real context streams}

Four public, ground-truthed streams test whether the distinct-item property exists outside the synthetic probe. The first three also appear in the companion mechanism paper; here they establish the kinds of context for which an inspectable slot table is natural.

\begin{table}[H]
\centering
\small
\caption{Bare allocate-on-novelty mechanism on real redundant streams.}
\begin{tabularx}{\textwidth}{L{0.14\textwidth} Y C{0.10\textwidth} C{0.10\textwidth} C{0.10\textwidth} Y}
\toprule
Domain & Stream & Events & Distinct & Redundancy & Cache outcome \\
\midrule
Systems logs & Loghub HDFS log lines & 2,000 & 14 & 143$\times$ & 15 templates; grouping accuracy 0.89; 133$\times$ compression \\
Clinical & MIMIC-IV demo prescription events & 18,087 & 631 & 29$\times$ & 628 slots; route recall 0.995 \\
Claims & DE-SynPUF inpatient diagnosis codes & 150,000 & 3,981 & 38$\times$ & 3,933 slots; chapter recall 0.989 \\
Mapping & KITTI odometry sequence 00, place cells & 4,541 & 357 & 13$\times$ & 357 slots; revisit recall 1.00 \\
\bottomrule
\end{tabularx}
\end{table}

On Loghub HDFS, whitespace-tokenized lines, positional match similarity, and wildcard generalization on merge turn the cache into an online log parser without further engineering. It recovers 15 templates against 14 ground-truth templates and groups lines at 0.89 accuracy, compressing 2,000 lines into 15 inspectable slots. This is not competitive with tuned production parsers on more diverse logs: on BGL, the bare token similarity reaches only 0.47, where engineered parsers do better. The point is that the cache recovers the parser family's core allocation loop, not that it displaces mature systems \citep{zhu2023}.

On the MIMIC-IV open demo, 18,087 prescription events yield 628 slots, one per distinct drug to within three, and nearest-slot reading recovers the administration route at 0.995, matching an attention memory that stores every event at 29 times the size \citep{johnson2023}. On DE-SynPUF, 150,000 inpatient diagnosis events yield 3,933 slots against 3,981 distinct codes and recall each code's ICD-9 chapter at 0.989; a fixed-budget eviction cache at one quarter of the distinct count collapses to 0.30 \citep{cms2010}.

The KITTI row makes the context interpretation particularly clear. Places are ten-metre occupied cells and the query is a revisit: a frame whose place was last seen at least 500 frames earlier. At a budget equal to the distinct-place count, the sliding window recalls none of the revisits on any of three sequences because every revisit exceeds the matched window. The heavy-hitter is unstable across sequences, from 0.34 to 1.00, because it retains popular places and drops the tail. Allocate-on-novelty recalls every revisit on every sequence while holding exactly one slot per place. The memory grows with the environment and remains flat along the trajectory, the invariance a mapping context should have \citep{geiger2012}.

Across these domains the retained context becomes a concrete object: template catalog, formulary, code set, or place map. The companion paper adds two mechanism-level boundaries relevant here. First, the threshold does not require per-domain tuning across the recommendation, clinical, and claims streams: any value on a broad plateau yields the same distinct-item cache. Second, on bursty BGL alerts, a surprise-adaptive budget recalls 0.94 at 18 average slots where a matched fixed budget recalls 0.75; on nearly stationary MovieLens windows it ties the matched fixed budget, as it should \citep{palrojkova2026}.

\section{A trained control study: summary plus distinct recall at half the cache}

Mechanism-level allocation does not establish what happens when the cache is embedded in a model trained end to end, where the representations themselves move under the gate. Character-level language modeling on enwik8 isolates the read without a domain frontend. We train four-layer residual stacks of width 256, sequence length 512, batch size 32, and 6,000 steps over the 90 MB training split, using identical schedules, and evaluate bits per character on held-out data \citep{mahoney2006}.

\begin{table}[H]
\centering
\small
\caption{Matched reads on enwik8. The headline runs are single-seed T4 measurements; the replicated results follow in the text.}
\begin{tabularx}{\textwidth}{Y C{0.14\textwidth} C{0.18\textwidth} C{0.16\textwidth}}
\toprule
Read & Valid bpc & Attended fraction & Training time (s) \\
\midrule
Diagonal state-space & 2.193 & -- & 672 \\
Full attention & 2.087 & 1.00 & 882 \\
Novelty-gated attention & 2.073 & 0.49 & 1,122 \\
State-space + full attention & 1.932 & 1.00 & 1,143 \\
\textbf{State-space + gated cache (component)} & \textbf{1.877} & \textbf{0.53} & 1,404 \\
State-space + full attention + gated cache & 1.836 & 1.00 (and 0.53 gated) & 1,880 \\
\bottomrule
\end{tabularx}
\end{table}

Four findings build on one another.

First, novelty-gated attention reaches 2.073 versus 2.087 for full attention, a 0.014-bit margin within expected run-to-run variation at this scale and therefore treated as parity rather than a win. It does so while admitting 49\% of tokens. The fixed-state path, with no item cache, trails both by about 0.11 bits per character. At this horizon, half the learned keys carry the content-addressed information the task uses.

Second, adding the state-space path to full attention reaches 1.932, improving by 0.155 bits over attention alone. The scale of this gain is an order of magnitude larger than the attention-versus-gated difference and is consistent with the two paths doing different work: the recurrence carries smooth positional and local structure, while attention carries addressable recall.

Third, composing those two results gives the proposed component. State-space plus gated cache reaches 1.877 at an attended fraction of 0.53. It is nominally 0.055 below state-space plus full attention, a single-seed margin we do not rely on. The central statement is that restricting the content-addressed path to the novelty-kept half of the tokens gives up nothing against coupling recurrence with full attention.

Fourth, retaining both full attention and the gated cache reaches 1.836. It saves no cache because full attention remains, so it is not a deployment candidate. Its role is evidential: if the gated cache were merely a cheaper approximation to attention, adding it beside full attention should contribute nothing. Instead it improves the coupled read by a further 0.096 bits. Organizing context by distinct content supplies an inductive view not automatically extracted by dense attention.

Three boundaries accompany the table. The models are small and far from the enwik8 state of the art; the comparison is internal to matched backbones, not a leaderboard claim. The headline entries are single-seed. And the saving is in cache and inference read cost, not in training clock time: the unoptimized novelty scan is quadratic and makes the gated read slower per step than attention (1,122 versus 882 seconds; 1,404 for the component).

The seed boundary has been retired. Retraining the five principal reads over three seeds on Apple-silicon hardware gives: state-space 2.191$\pm$0.016, full attention 2.029$\pm$0.064, gated read 2.014$\pm$0.065 at fraction 0.49, coupled read 1.933$\pm$0.009, and component 1.922$\pm$0.035 at fraction 0.51. Across ten contiguous validation blocks per seed, the gated read is at or below full attention in 29 of 30 seed-block pairs. The component is below full attention in all 30. Against the full-attention coupled read, the component wins all ten blocks on two seeds and loses all ten on the third. Thus the durable result remains parity with the full-attention coupling at half the attended tokens; any further margin is within seed variation.

A selective, input-gated recurrence changes the best text result. Paired with the gated cache it reaches 1.709$\pm$0.008 bpc at an admitted fraction of 0.61 over three seeds, two tenths below the plain component with every seed separated, below the three-path read's 1.836, and with the tightest spread in the study. On concatenated source code the same coupling reaches 1.394$\pm$0.019 at fraction 0.50, four tenths below its cache-free ablation (1.811$\pm$0.008) and about half a bit below full attention, whose seeds vary by $\pm$0.28. This is the strongest text coupling measured. It is also sharply domain-specific: on the claims task below it collapses, showing that the recurrent summary operator must match the stream even when the novelty cache remains useful.

\section{Context scaling: the advantage grows with horizon}

The decisive question for long-context engineering is how the gap changes with horizon. We sweep sequence length over 256, 512, and 1,024 tokens, adjust batch size to 24, 16, and 8 to remain feasible, and train full attention, novelty-gated attention, and the component for 3,000 steps at each horizon. Within each horizon the reads share the same training budget, so those gaps are clean. Across horizons, the fixed step count means longer sequences receive a different optimization budget; absolute loss trends across horizons are not the claim. These runs are also shorter than the 6,000-step control above.

\begin{table}[H]
\centering
\small
\caption{Context-length sweep, three seeds per cell.}
\begin{tabularx}{0.96\textwidth}{C{0.12\textwidth} C{0.19\textwidth} C{0.19\textwidth} C{0.19\textwidth} Y}
\toprule
Horizon $T$ & Full attention & Novelty-gated & \textbf{Component (ours)} & Attention $-$ component \\
\midrule
256 & 2.359$\pm$0.018 & 2.368$\pm$0.023 & \textbf{2.419$\pm$0.032} & $-0.060$ \\
512 & 2.457$\pm$0.012 & 2.449$\pm$0.013 & \textbf{2.406$\pm$0.067} & $+0.051$ (within spread) \\
1,024 & 2.693$\pm$0.029 & 2.667$\pm$0.030 & \textbf{2.393$\pm$0.054} & $+0.300$ \\
\bottomrule
\end{tabularx}
\end{table}

The component remains nearly flat as the horizon quadruples (2.42, 2.41, 2.39), while matched attention degrades (2.36, 2.46, 2.69). The gap moves from $-0.060$ at 256, through a favorable but non-claimable $+0.051$ within spread at 512, to $+0.300$ at 1,024 with cleanly separated spreads. The novelty-gated read tracks attention at each horizon, so the parity result is not confined to 512 tokens. The component's admitted fraction falls from approximately 0.45 to 0.42 as context lengthens: the longer context accumulates more redundancy, and the memory saving grows where the performance advantage does.

The short-horizon reversal is part of the result. At 256 tokens the relevant context fits comfortably in attention and the recurrent path contributes no measurable benefit. The component is a long-context design, not a universal replacement for short attention.

\begin{figure}[H]
\centering
\includegraphics[width=0.68\textwidth]{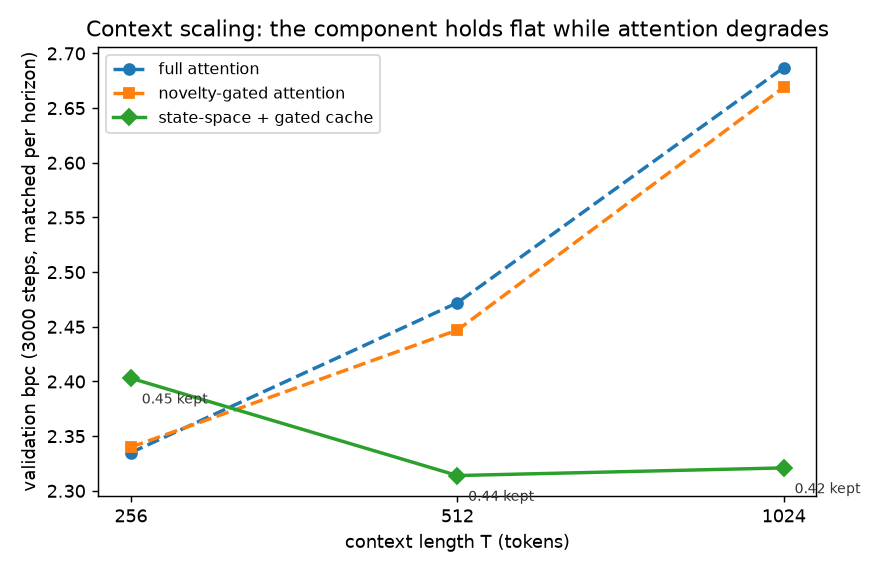}
\caption{Context scaling at matched budget per horizon. The component (state-space plus gated cache) remains nearly flat from 256 to 1,024 tokens while full attention and the gated read alone degrade. The annotations report the component's admitted fraction, which falls as the horizon grows.}
\end{figure}

\section{Which context to keep depends on how the task uses the past}

A frozen-threshold sweep traces the trained cache-size frontier; a sliding-window read at matched budgets determines what that frontier means. The result is deliberately reported against a novelty-first interpretation.

Thinning helps both policies on short-horizon character prediction. The gated read improves monotonically as its admitted fraction falls from 0.75 to 0.20, from 2.350 to 2.305 bpc at 3,000 steps. The window improves as it narrows, and the smallest window is the best point in the sweep: 2.262 at a 0.12 fraction. Near a matched fraction of 0.2, the policies are at parity: 2.305 at 0.20 for novelty versus 2.312 at 0.22 for the window. The task is locality-carried, so recency is the correct context policy.

\begin{table}[H]
\centering
\small
\caption{Frozen novelty thresholds against sliding windows on the control corpus. ``Blocked'' reports the blocked evaluation estimate used in the original study.}
\begin{tabular}{ccc|ccc}
\toprule
\multicolumn{3}{c|}{Novelty-gated (frozen $\tau$)} & \multicolumn{3}{c}{Sliding window}\\
Fraction & bpc & Blocked & Fraction & bpc & Blocked\\
\midrule
0.75 & 2.350 & 2.343$\pm$0.034 & 0.47 & 2.337 & 2.329$\pm$0.034\\
0.52 & 2.332 & 2.324$\pm$0.033 & 0.38 & 2.327 & 2.318$\pm$0.033\\
0.29 & 2.324 & 2.317$\pm$0.034 & 0.22 & 2.312 & 2.304$\pm$0.034\\
0.20 & 2.305 & 2.298$\pm$0.033 & 0.12 & 2.262 & 2.258$\pm$0.033\\
\bottomrule
\end{tabular}
\end{table}

Training the deployed budget policies end to end reveals a second boundary. Sink-plus-window trains as stably as attention and the SnapKV-style read nearly so, but the heavy-hitter read trains poorly: 2.69 and 3.14 bpc at its two budgets versus 2.09 for attention. Selection by accumulated attention mass is noisy while the attention itself is untrained, and the hard selection feeds the noise back into learning. Heavy-hitter eviction is a plausible inference-time policy; it is a poor training-time primitive in this experiment. Novelty does not depend on the model's own accumulated saliency statistics and is therefore better behaved during learning.

Two corpus controls narrow the interpretation further. Source code is substantially more compressible than prose by the attention baseline (1.62 versus 2.35 bpc), yet the learned gate settles near half on both (0.47 prose and 0.50 code). The admitted fraction does not measure corpus-level compressibility; it reflects local similarity in the learned key space. On code, the gated read has a small deficit to attention (1.640 versus 1.622), whereas on prose it has a small surplus. A selective recurrence without attention reaches 1.912 at the 6,000-step budget, within the coupled reads' seed band, reinforcing that this short control corpus does not require strong item-addressed recall.

The sweep also asks which occurrence of a repeated key should be kept. Allocate-on-novelty keeps the first. Its mirror keeps only the latest occurrence of each group of similar earlier keys, matching the copying behavior associated with induction heads \citep{olsson2022}. Under the same protocol, keep-latest reaches 2.267$\pm$0.002 bpc at an admitted fraction of 0.21 over three seeds. It beats every keep-first threshold point at less than half their fraction, beats the matched 0.22 sliding window (2.312), and is statistically tied with the narrowest window's 2.262 at fraction 0.12. On source code it is the steadiest read measured, 1.731$\pm$0.043 versus 1.94$\pm$0.28 for full attention and keep-first at that budget.

This revises the trained read, not the mechanism. At the mechanism level a slot represents the whole cluster, and the companion paper's merge study found the stored occurrence immaterial to the result. In a trained attention read, however, the retained value carries positional context, and the most recent occurrence can be the useful one. That result is again domain-dependent: on the thousand-event claims task, keep-latest scores 8.55$\pm$0.17, behind keep-first at 8.48$\pm$0.08, behind full attention, and well behind the component. Recency wins when the repeated occurrence's latest local context matters; first-allocation wins when the slot is intended to represent a stable distinct entity.

\section{A recall-carried long-context task: the crossover on real claims}

The control corpus identifies boundaries but does not force distinct-entity recall. The end-to-end structured task is next-code prediction on DE-SynPUF inpatient claims. For each member, claims are ordered by date and represented as a member boundary followed by, for each claim, a claim boundary and its ICD-9 diagnosis codes. The model predicts the next token. Evaluation uses bits per event and top-1 accuracy on held-out members, so generalization is across patients rather than within a patient's history. The stream contains 576,885 events and a vocabulary of 4,096 codes and boundaries. Four reads are trained for 3,000 steps at three horizons, with the batch scaled to hold the token budget fixed at the two longer horizons.

\begin{table}[H]
\centering
\scriptsize
\caption{DE-SynPUF next-code prediction, three seeds per cell. Parentheses report admitted fractions.}
\begin{tabularx}{\textwidth}{C{0.12\textwidth} C{0.19\textwidth} C{0.22\textwidth} C{0.18\textwidth} Y}
\toprule
Horizon & Full attention & Novelty-gated & Coupled & \textbf{Component (ours)} \\
\midrule
256 & 7.991$\pm$0.092 & 7.947$\pm$0.047 (0.49) & 7.986$\pm$0.016 & \textbf{7.969$\pm$0.023 (0.50)} \\
512 & 8.078$\pm$0.060 & 8.067$\pm$0.096 (0.48) & 8.046$\pm$0.032 & \textbf{8.077$\pm$0.029 (0.47)} \\
1,024 & 8.396$\pm$0.041 & 8.483$\pm$0.077 (0.53) & 8.107$\pm$0.049 & \textbf{8.085$\pm$0.040 (0.45)} \\
\bottomrule
\end{tabularx}

\vspace{0.3em}
\parbox{0.97\textwidth}{\footnotesize Deployed policies at horizon 1,024 and budget 256: heavy-hitter 8.419$\pm$0.005; SnapKV-style 8.584$\pm$0.020; sliding window 8.633$\pm$0.107; sink-plus-window 8.628$\pm$0.105, three seeds each.}
\end{table}

At 256 and 512 events, all four reads tie within seed spread. At 1,024 events the ordering separates. The component reaches 8.085$\pm$0.040 bits per event, 0.311 below full attention with non-overlapping spreads, while attending over 45\% of events; the plain coupled read follows at 8.107. The crossover between 512 and 1,024 events matches the shape of the enwik8 horizon sweep. An event contains more semantic content than a character, and the member history relevant to the next diagnosis spans hundreds of events.

The matched-budget policies determine which context rule is correct. The sliding window that led the short character sweep trails the component by more than half a bit. The strongest deployed policy, the heavy hitter, trails by about a third. The rare and distinct codes matter, so popularity and recency are the wrong retention criteria. The fixed-state backbone at the short horizon scores 8.021, confirming the bounded-state floor. As in the context sweep, only within-horizon comparisons are claimed because the training step count is fixed across horizons.

The selective coupling that leads prose and code fails severely here. At the 1,024-event horizon, the selective backbone alone reaches 9.64$\pm$0.07 and with the cache 9.77$\pm$0.05, approximately 1.5 bits behind the plain component. The input-gated recurrence that suits smooth text streams fails on the spiky categorical stream, and the cache cannot rescue a failed summary path. There is no universal recurrent companion: selective summary wins text; plain linear summary wins claims; keep-latest wins text budgets and loses claims. What is stable across the three corpora is narrower: the best measured read on prose, code, and claims contains the novelty-gated cache, while the domain determines the other path.

\section{The rest of the program, measured}

\subsection{Incremental cost}

At inference, the gate is incremental. The cache contains only admitted keys; each arriving key is compared with the occupied set; a query reads over the occupied set alone. Streaming 65,536 tokens of width 256 at eightfold redundancy on one CPU, the cache answers a query in 64 microseconds, while attention over the full stream takes 1,355 microseconds: a 21-fold read speedup with an eightfold smaller cache. The cost is allocation: 56 microseconds per incoming token versus 0.3 microseconds for attention's append. This reproduces at scale the companion paper's build-versus-read law. The cache is favorable when the retained context is queried repeatedly or memory is binding, not when a sequence is built once and read once.

\subsection{Operational logs at full scale: one clean result and one wall}

The trained log study uses fully annotated Loghub-2.0 releases under a strict temporal split: the first 90\% of each stream for training and a later contiguous slice for evaluation, three seeds per read. Copy-last-template accuracy is reported as the free floor supplied by a storm-dominated log.

On HDFS, 1.8 million lines and 37 templates, copy-last accuracy is 0.574. Every learned read is far below the uniform loss floor, approximately 0.98 bits per event versus 5.21, and above the copy floor at 0.66 accuracy. The gated read matches full attention at cache fraction 0.42. The component leads attention on all three seeds, 0.975$\pm$0.008 versus 0.990$\pm$0.011 bits, at fraction 0.39. The selective coupling ties the plain component at 0.974$\pm$0.006 and fraction 0.40, showing that its failure on claims is not a universal failure on event data. Keep-latest defines an extreme-memory point: repeated storm lines make the latest duplicate sufficient, reducing the fraction to 0.03--0.04, roughly one tenth of the other reads, at a cost of about 0.09 bits (1.08$\pm$0.06 versus 0.975). It is a frontier point, not the winner.

On BGL, 4.6 million lines and 320 templates, every read fails under the same protocol. Losses range from 9.5 to 12.5 bits versus an 8.3 uniform floor, accuracy lies far below the copy floor, and seed spreads reach a bit or more. The later slice changes regime: BGL's alert-template distribution drifts over months. A model confidently wrong on the future slice has encountered distribution shift, not a memory-capacity limit. The trained log claim is therefore restricted to streams whose template distribution is sufficiently stationary across the split. Rolling refits or within-regime evaluation are required for drifting streams.

\subsection{One stream, two targets: recall-carried versus summary-carried}

The original claims cost head predicted the payment decile of the arriving claim and produced validation losses above the ten-class uniform floor for every read, indicating memorization of training members. That experiment was withdrawn. The redesign adds outpatient claims, increasing the data by approximately twelvefold, plus dropout, weight decay, and best-checkpoint selection on a held-out slice.

The redesigned task clears the 3.322-bit uniform floor. Every read transfers at 3.268--3.276 bits per claim, with three-seed spreads of $\pm$0.003. The ordering, though modest, is diagnostic: the two recurrent-path reads tie at 3.268 for the summary alone and for the component, while full attention and the gated read tie at 3.276. The cache neither helps nor hurts; the recurrent summary carries the signal. The extensions obey the same split to the third decimal: selective recurrence and its coupling reach 3.269--3.270, and keep-latest joins the attention side at 3.278.

Thus the same underlying stream supports two different context requirements. Next-code prediction is recall-carried and favors the distinct-item cache. Cost forecasting is summary-carried and favors recurrence. Context policy cannot be selected from the corpus alone; it must be selected from the task's dependency on the past.

\subsection{The trained adaptive budget: a settled negative in this setting}

Inside a trained model, the surprise-modulated threshold ties the fixed gate to three decimal places on both standard and deliberately bursty enwik8 while admitting more of the stream. The first burst construction had phases longer than the training window, leaving each window quasi-stationary. A repaired test uses bursts four times shorter than the window so every window crosses several regime changes. The result remains a tie at added cost: 1.137 versus 1.137 bpc, admitted fraction 0.71 versus 0.49.

The interpretation is that the learned soft gate already adapts to content token by token. An explicit budget schedule layered on top is redundant inside these windowed trained models. The adaptive concentration remains useful at the mechanism level for truly streaming caches with phase-level demand swings, as shown by BGL alert storms in the companion study, but is not justified by the present trained backbones.

\subsection{Architecture robustness}

Widening the four-layer backbone preserves the ordering. At width 384, the component reaches 1.839, the full-attention coupled read 1.849, and attention 1.945. At width 512, attention reaches 1.894, the component 1.823, and the selective coupling 1.686 at admitted fraction 0.63. The width axis is therefore covered at 256, 384, and 512 with the same qualitative result.

At six layers and width 256, the initial 5,000-step experiment inverts: attention and the gated read lead at 1.91--1.92 and the coupled reads trail by approximately 0.08. A plausible interpretation is that additional attention layers learn local composition that substitutes for the recurrent path while the coupled optimization cost remains. Doubling the training budget to 10,000 steps changes the conclusion: the component reaches 1.707 at fraction 0.59 versus 1.731 for attention, with blocked means 1.747$\pm$0.034 versus 1.774$\pm$0.035 in the same direction. The earlier inversion was an optimization artifact. At depth, the coupling requires more training budget.

\section{Context-engineering consequences}

The experiments support a design rule more specific than generic prompt compression.

\subsection{Three memory structures, three dependency types}

A practical long-context system should not force every kind of information through one memory policy.

\begin{table}[H]
\centering
\small
\caption{The context policy suggested by the measured dependency on the past.}
\begin{tabularx}{0.94\textwidth}{L{0.22\textwidth} Y Y}
\toprule
Dependency type & Appropriate memory & Evidence in this paper \\
\midrule
Locality-carried & Sliding or recency window & Short-horizon character prediction: the narrowest window reaches 2.262 bpc and matches or beats novelty at lower fraction. \\
Summary-carried & Recurrent/state-space summary & Claims cost forecasting: recurrent reads lead at 3.268 while attention and gated cache reach 3.276. \\
Recall-carried & Novelty-gated distinct-item cache & Claims next-code prediction, HDFS templates, clinical codes, place revisits, and the companion associative-recall studies. \\
Mixed & Coupled summary + distinct-item cache & enwik8 and claims at long horizons; the component matches or exceeds full-attention coupling at approximately half the attended tokens. \\
\bottomrule
\end{tabularx}
\end{table}

This decomposition also explains why a single benchmark can be misleading. A locality-dominated corpus makes a window look universally sufficient; a recall probe makes novelty look universally superior; an aggregate target makes recurrence look sufficient. The architecture should be judged on a set of tasks that deliberately separates these dependencies.

\subsection{Context objects, not merely compressed tokens}

The strongest use of the cache is not to replace a tokenizer with clustering. It is to maintain a set of addressable context objects whose size follows the distinct information active in the stream. In a general context engine, a slot should therefore carry more than a hidden key--value pair: source references, timestamps or offsets, merged evidence, confidence, and an update history. Those additions do not change the measured mechanism, but they turn its inspectable slot set into an auditable working memory.

The paper does not establish that arbitrary language turns will organize themselves into stable human-readable facts. The real-stream alignments arise because the frontends bind keys to explicit movies, templates, drugs, codes, and places. Applying the method to retrieval-augmented generation or multi-turn assistants requires a frontend and merge semantics that preserve propositions, entities, negation, temporal updates, and source authority. A novelty gate can decide that two representations are close; it cannot by itself decide whether one statement supersedes another or whether two similar passages conflict.

\subsection{The relevant scaling law}

Let $N$ be observed context units and $K$ the number of distinct retained items. Full attention stores $O(N)$ keys and reads $O(N)$ per query. The present cache stores approximately $O(K)$ and reads $O(K)$, while naive online allocation costs $O(NK)$ because each incoming key scans occupied slots. When redundancy is high, $K\ll N$; when every item is novel, $K\approx N$ and the cache offers no asymptotic saving. Approximate nearest-neighbour indexing, hierarchical slots, or batched allocation are engineering requirements for scaling the write path, not changes to the principle.

The measured incremental result makes the trade concrete: at $N=65{,}536$ and eightfold redundancy, reading is 21 times faster and the cache is eight times smaller, while each write is much more expensive than attention's append. The right deployment is persistent, repeatedly queried context, not transient one-shot prompting.

\section{Scope}

All experiments use public data: Loghub, the MIMIC-IV open demo, CMS DE-SynPUF synthetic claims, KITTI, and enwik8. No proprietary stream, schema, or domain knowledge from an operational deployment enters the experiments. Mechanism-level studies use synthetic keys bound to real entity streams and train-free reads; they validate cache statistics and, for logs, one ground-truthed end task, but are not trained domain models. The trained control, claims task, and context-length sweep are replicated over three seeds, with paired per-block statistics for the control. The corpus-swap and some architecture-robustness measurements are single-seed where stated. Compared runs share hardware; the headline control table uses an NVIDIA T4, while seed replicates, context sweeps, claims, and mechanism studies use Apple M1 and M4 systems.

The evidence does not yet include long multi-turn conversations, retrieval-augmented generation, tool-using agents, or production context builders. The paper therefore does not claim that the current cosine gate and merge rule are sufficient for semantic facts in those systems. It establishes a lower-level result: a learned novelty-gated content-addressed path can retain distinct information at substantially below token-count cost, can be coupled with recurrent summaries without surrendering performance, and has measurable regimes where it is preferable to recency or fixed-budget eviction. Extending the frontend to proposition-level context, preserving contradictions and temporal updates, evaluating answer faithfulness and provenance, and optimizing the allocation scan are the next steps.

Within those limits, the central context-engineering claim is deliberate: context need not be a transcript, a fixed summary, or a window. It can be a structured working set whose size follows the distinct information the task may need to recall, while recurrence carries aggregates and recency carries local texture.

\section*{Acknowledgements}
The seed replicates, the claims task, and the mechanism-level studies were designed, run, and verified on Apple silicon (an M1 and an M4 machine), and we are grateful to Google Colab for access to the NVIDIA T4 GPUs, on which the headline control study and the context sweep were run. We thank Aarav Pal for his help with the data downloads and with testing and running the experiments.

\end{document}